\pdfoutput=1
\documentclass[11pt]{article}

\usepackage[preprint]{acl}

\usepackage{algorithm}
\usepackage{algorithmicx}
\usepackage{algpseudocode}
\usepackage{booktabs}
\usepackage{multirow}

\usepackage{times}
\usepackage{latexsym}
\usepackage{amsmath}
\usepackage[T1]{fontenc}
\usepackage[utf8]{inputenc}

\usepackage{microtype}

\usepackage{inconsolata}

\usepackage{graphicx}

\title{FAEDKV: Infinite-Window Fourier Transform for Unbiased KV Cache Compression}

\author{Runchao~Li, \, Yao~Fu, \, Mu~Sheng, \, Xianxuan~ Long, \, Haotian~Yu, \, Pan~Li \\ Case Western Reserve University}
\begin{document}
\maketitle
\begin{abstract}
The efficacy of Large Language Models (LLMs) in long-context tasks is often hampered by the substantial memory footprint and computational demands of the Key-Value (KV) cache. Current compression strategies, including token eviction and learned projections, frequently lead to biased representations—either by overemphasizing recent/high-attention tokens or by repeatedly degrading information from earlier context—and may require costly model retraining. We present FAEDKV (Frequency‐Adaptive Infinit\textbf{e}-Win\textbf{d}ow for KV cache), a novel, training-free KV cache compression framework that ensures unbiased information retention. FAEDKV operates by transforming the KV cache into the frequency domain using a proposed Infinite-Window Fourier Transform (IWDFT). This approach allows for the equalized contribution of all tokens to the compressed representation, effectively preserving both early and recent contextual information. A preliminary frequency ablation study identifies critical spectral components for layer-wise, targeted compression. Experiments on LongBench benchmark demonstrate FAEDKV's superiority over existing methods by up to \textbf{22\%}. In addition, our method shows superior, position-agnostic retrieval accuracy on the Needle-In-A-Haystack task compared to compression based approaches.
\end{abstract}
\begin{figure*}[t]
  \includegraphics[width=2\columnwidth]{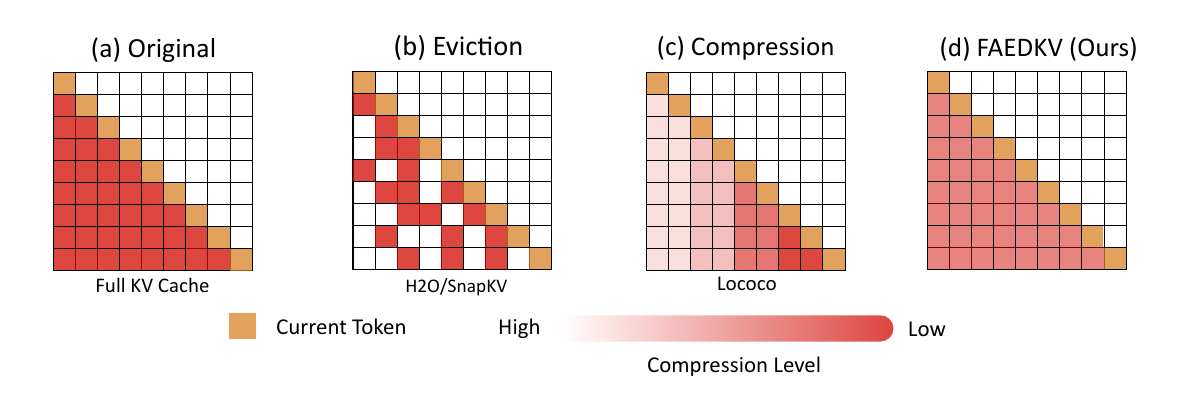}
  \caption{Conceptual illustration of how different KV cache management strategies process past tokens, highlighting their inherent biases. (a) Full KV Cache: Represents unbiased, complete retention. (b) Eviction strategies (e.g., H2O/SnapKV): Clearly show biased token removal. (c) Learned compression (e.g., Lococo): Illustrates bias through targeted, often heavier, compression of older tokens. (d) FAEDKV (Ours): Its visually consistent treatment of all past tokens underscores an algorithmic approach designed to operate without introducing arbitrary biases.}
  \label{fig:intro}
\end{figure*}
\section{Introduction}

LLM has become the paradigm in language-generating tasks. It can perform variety of important tasks such as text generation, question answering, mathmatical problem-solving. For this types of problems, a long context is often required to provide enough background information, thus they demand the model have long context capability. Recently, chain-of-thought reasoning models have earned popularity due to their ability to breakdown the problems into small steps and solve them in a reasoning process. It also requires sufficiently long generated text to solve complex problems.

However, Transformers\cite{vaswani2023attentionneed} inherently struggle with long sequences. While their quadratic self-attention complexity is a known bottleneck, autoregressive decoding mitigates this for subsequent tokens using a Key-Value (KV) cache. This cache, however, introduces its own challenge: its memory footprint grows linearly with context length, rapidly consuming inference resources.

Recent approaches to mitigate KV cache memory costs primarily involve token pruning. For instance, H2O \cite{zhang2023h2oheavyhitteroracleefficient} evicts tokens with low accumulated attention scores to maintain a target cache size, while PyramidKV \cite{cai2024pyramidkvdynamickvcache} extends this by dynamically allocating cache budgets across layers and selecting tokens deemed most important. While such methods can reduce KV cache sizes during inference, their reliance on attention scores as a primary selection criterion introduces a bias. This bias favors tokens with high immediate relevance to the current query, potentially leading to the premature eviction of important tokens, a phenomenon related to the 'lost in the middle' problem \cite{liu2023lostmiddlelanguagemodels}.

Alternatively, learning-based compression techniques have been applied to the KV cache. For example, ActivationBeacon \cite{zhang2024longcontextcompressionactivation} learns to condense preceding tokens into a compact 'activation beacon,' while LOCOCO \cite{cai2024lococodroppingconvolutionslong} employs 1-D convolutional networks to project keys and values into compressed representations. Although these data-driven approaches can effectively reduce cache size, they often rely on a learned compression module activated when the cache exceeds a predefined threshold. This can lead to repeated compression of earlier tokens as the context window expands, progressively degrading their information content. Moreover, these methods frequently necessitate model fine-tuning or the training of auxiliary modules, demanding significant computational resources.

This paper introduces Frequency‐Adaptive Infinite Window for KV cache (FAEDKV), a novel algorithm to address these challenges. FAEDKV transforms the KV cache into the frequency domain, ensuring balanced preservation of information from all tokens. Its methodology involves a layer-wise frequency ablation study to identify critical spectral components and a novel Infinite-Window Fourier Transform (IWDFT) for managing the frequency-domain cache. Unlike common approaches leading to abrupt eviction or repeated compression of older tokens (conceptualized in Figure \ref{fig:intro} (b,c)), FAEDKV more consistently retains their information (Figure \ref{fig:intro} (d)), enabling targeted frequency filtering for compression. FAEDKV is universally compatible, requires no fine-tuning, and integrates via a one-time ablation study and modification of the attention module.

% Our method adapt to both pre-filling and inference efficiently. 

Contribution:

\begin{itemize}
    \item We propose \textbf{FAEDKV}, a novel method achieving \textbf{unbiased Key-Value (KV) cache compression} by transforming entries into the frequency domain, thereby mitigating prevalent contextual biases found in existing techniques.
    \item We develop a supporting frequency-based framework featuring a novel \textbf{Infinite-Window Fourier Transform (IWDFT)} for efficient, recursive cache updates, and a \textbf{frequency ablation study} for targeted, layer-wise spectral pruning to optimize compression.
    \item Experiments demonstrate FAEDKV's superior performance, significantly outperforming established baselines on LongBench by up to 22\% with 9\% cache size and achieving consistent, position-agnostic retrieval accuracy in Needle-in-a-Haystack tests.
\end{itemize}

\begin{figure*}[t]
  \includegraphics[width=2\columnwidth]{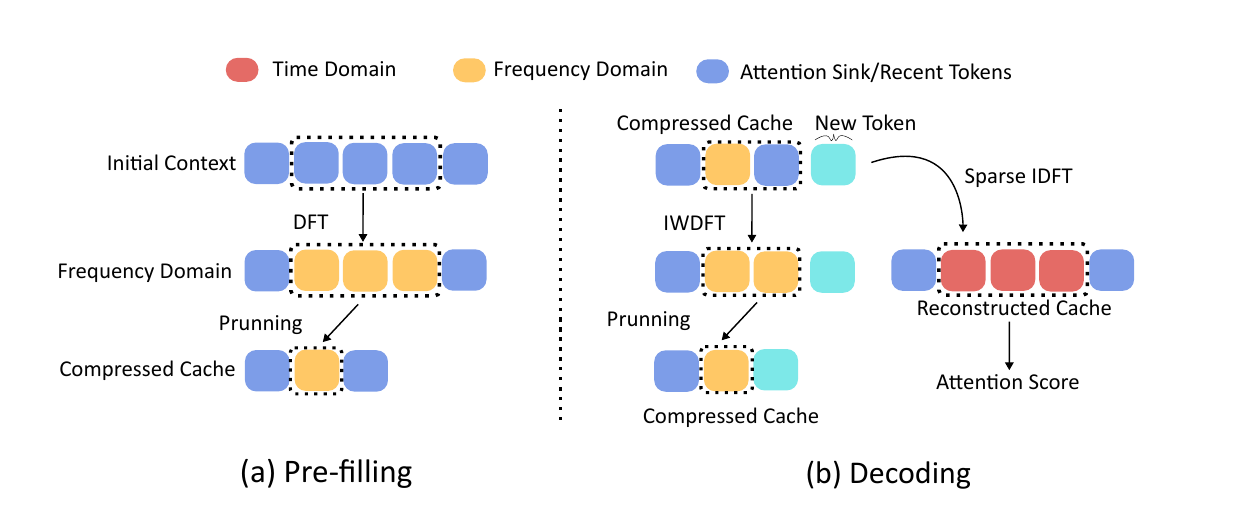}
  \caption{Overview of the FAEDKV workflow. (a) Pre-filling: The middle segment of the initial context is converted to the frequency domain (orange) by DFT, pruned, and stored with sink/recent tokens (blue). (b) Decoding: The compressed frequency-domain segment (orange) is updated via IWDFT (with new token information) and re-pruned. For attention, it's reconstructed to the time domain (red) by Sparse IDFT and combined with sink/recent tokens (blue) before attention score calculation.}
  \label{fig:workflow}
\end{figure*}

\section{Related Works}
\subsection{KV Cache Compression}
\label{sec:appendix}
Managing the extensive Key-Value (KV) cache in Large Language Models (LLMs) to reduce memory overhead and latency is a significant research focus \cite{ge2023surveykvcache, liu2024deja}. Many approaches selectively prune the cache by evicting less important tokens based on attention scores or other heuristics, such as H$_2$O \cite{zhang2023h2o}, Scissorhands \cite{liu2023scissorhands}, and SnapKV \cite{li2024snapkvllmknowslooking}. Other strategies involve learned projections or recursive compression, like LoCoCo \cite{cai2024lococodroppingconvolutionslong}, which may require fine-tuning as discussed in broader analyses of learned long-context methods \cite{tan2024llocolearninglongcontexts}. Architectural modifications like Grouped-Query Attention (GQA) \cite{ainslie-etal-2023-gqa} or other attention-aware compression techniques \cite{ge2024modeltellsdiscardadaptive} also aim to reduce cache size. While these methods effectively decrease memory usage, they often result in an unequal influence of tokens on the compressed cache, with some tokens being discarded or heavily down-weighted. Our work diverges by aiming to ensure a more balanced contribution from all tokens to the compressed representation.
\subsection{Fourier Transform in LLM}
Concurrently, Fourier transforms are proving instrumental in analyzing and enhancing LLMs. Studies reveal LLMs utilize Fourier features internally for tasks like arithmetic, encoding information across different frequency components \cite{nanda2023emergent, Murty2024ExactlyST}. Beyond analysis, frequency-domain techniques are actively improving model efficiency and capabilities. For example, Transformer FFNs have been reinterpreted as frequency transformers \cite{lee2024scaling}, and methods like AFFormer \cite{li2024afformer} incorporate Fourier-inspired adaptive filters. Fourier features are also applied in specialized areas such as LLM-based time-series forecasting \cite{zhou2024sfm}, creating robust positional embeddings like RoFormer \cite{su2024roformer}, and developing novel frequency-domain attention mechanisms like FNet \cite{lee2021fnet}. While these studies highlight the versatility of Fourier methods in LLMs, their specific application to KV cache compression with an emphasis on uniform token contribution---as explored in our work---remains a novel direction.

\section{Preliminaries}
Existing KV cache compression largely relies on eviction or learning-based techniques, which can introduce recency bias. Our approach differs by applying a novel frequency-domain transformation to the KV cache, aiming for equalize token contribution to the compressed state. This requires empirically analyzing frequency component importance via ablation experiments that measure perplexity changes upon pruning. This section outlines preliminaries and the frequency ablation study.

\subsection{Background}
\subsubsection{KV Cache in Autoregressive Decoding}

During generation, we omit batch and head dimensions for clarity.  Let 
\(\mathbf{x}^{t}\in\mathbf{R}^{1\times d}\) be the input embedding at step \(t\), and let 
\(W^Q, W^K, W^V\in\mathbf{R}^{d\times d}\) be the projection matrices.  The query, key, and value vectors are computed as
\begin{equation}
\mathbf{q}^{t} = \mathbf{x}^{t} W^ Q, 
\mathbf{k}^{t}= \mathbf{x}^{t} W^ K, 
\mathbf{v}^{t}= \mathbf{x}^{t} W^ V.
\end{equation}
The key and value caches grow by appending the new vectors:
\begin{equation}
\mathbf{K}^{t+1}= 
\begin{bmatrix}
\mathbf{K}^{t} \\[4pt]
 \mathbf{k}^{t}
\end{bmatrix},
\mathbf{V}^{t+1} = 
\begin{bmatrix}
\mathbf{V}^{t} \\[4pt]
 \mathbf{v}^{t}
\end{bmatrix},
\end{equation}
so that \(K^{t},V^{t}\in\mathbf{R}^{t\times d}\).

The attention output for the next token is then
\begin{equation}
\label{eq:kv-attn}
o^{t}
= \mathrm{Softmax}\!\Bigl(\frac{\mathbf{q}^{t}\,\bigl[ \mathbf{K}^{t}\bigr]^\top}{\sqrt{d}}\Bigr)\, \mathbf{V}^{t}.
\end{equation}

While caching prevents the redundant recomputation of past keys and values, allowing the attention calculation for each new token to be performed in $O(t)$ time with respect to the current context length t, the KV cache itself incurs a memory cost of $O(t)$. For very long sequences, both this linear growth in memory and the per-step computational cost become substantial. This challenge motivates our frequency-domain compression strategy, aimed at reducing the effective cache size—and thereby both memory and computational overheads for attention—without sacrificing the model’s ability to attend to both recent and distant tokens.

\subsubsection{Discrete Fourier Transform}
\label{sec:dft_preliminaries} % Optional: if you want to reference this subsection

The Discrete Fourier Transform (DFT) is a fundamental tool for analyzing signals in the frequency domain, widely used in fields like voice and image processing. In the context of Large Language Models (LLMs), a sequence of vectors (such as those in the KV cache along the token dimension) can be treated as a 1-D time-domain signal. Applying a 1-D DFT to such sequences transforms them into the frequency domain, which can reveal structural properties and allow for targeted manipulation, forming the basis for our compression approach.

The DFT converts a finite discrete-time sequence $x[0], x[1], \dots, x[N-1]$ into its frequency-domain representation $X^f[0], X^f[1], \dots, X^f[N-1]$. This transformation is defined as:
\begin{equation}
\label{eq:dft-full}
\begin{aligned}
    X^f[k] &= \sum_{n=0}^{N-1} x[n] (W_k)^n, \\
    \text{where } W_k &= e^{-j\frac{2\pi k}{N}}, \quad k=0, 1, \dots, N-1.
\end{aligned}
\end{equation}
In this formulation, $x[n]$ is the input signal at time $n$, $X^f[k]$ is the $k$-th frequency component, $N$ is the sequence length, $j$ is the imaginary unit, and $W_k$ is the complex exponential term $e^{-j\frac{2\pi k}{N}}$ (often referred to as a twiddle factor). This transformation can be computed efficiently using the Fast Fourier Transform (FFT) algorithm, which has an $O(N\log N)$ time complexity and $O(N)$ memory cost.

To utilize the frequency-domain representation for attention, the time-domain KV cache vectors must be reconstructed. This is achieved using the \emph{inverse DFT} (IDFT). Given the frequency components $X^f[k]$ and our previously defined $W_k = e^{-j\frac{2\pi k}{N}}$, the IDFT reconstructs the time-domain sequence $\widetilde{x}[n]$ as:
\begin{equation}
\label{eq:inverse-swdft}
\begin{aligned}
    \widetilde{x}[n] &= \frac{1}{N} \sum_{k=0}^{N-1} X^f[k] (W_k^*)^n, \\
    \text{where } W_k^* &= e^{j\frac{2\pi k}{N}}, n=0, 1, \dots, N-1.
\end{aligned}
\end{equation}
This reconstruction can be performed efficiently using an Inverse Fast Fourier Transform (IFFT) with $O(N\log N)$ time complexity. We adopt this method for reconstructing our KV cache from its compressed frequency-domain representation. While the initial pre-filling stage for attention calculation over $N$ tokens is typically $O(N^2)$, our DFT-based enables us to reduce the initial latency. We verify it at Section~\ref{exp:effi}.

\subsection{Frequency Ablation Study}
\label{subsec:freq-ablation}
Previous studies \cite{He_2023,kai2025freqkvfrequencydomainkeyvalue} have employed the Discrete Cosine Transform (DCT) for analyzing model components, often finding energy concentrated in lower frequencies and thus retaining only these for compression. We choose a DFT approach over DCT because DCT's implicit symmetric signal extension (mirroring) contributes to its strong emphasis on lower-frequency bins. It could leads to the loss of critical higher-frequency details.  

To assess the relative importance of different spectral bands in the KV cache, we perform a layer-wise frequency ablation study. We randomly sampled 100 texts from WikiText-103-v1 \cite{merity2016pointer}, processing each up to the model's maximum token sequence length, denoted as $N$. For each Transformer layer $\ell \in \{1, \dots, L_{layers}\}$ (where $L_{layers}$ is the total number of layers), we compute the DFT of the keys and values along the token sequence length dimension. This DFT yields $N$ frequency bins. These $N$ bins are partitioned into $C$ contiguous \emph{chunks}. Each chunk $c \in \{1, \dots, C\}$ consists of $N/C$ frequency bins. Let $B_c \subset \{0, \dots, N-1\}$ denote the set of frequency indices belonging to chunk $c$.

During the ablation study, for each layer $\ell$ and each chunk $c$, we zero out (prune) all DFT coefficients $X^f_\ell[k]$ where $k \in B_c$. The modified coefficients $\hat{X}^f_\ell[k]$ are thus defined as:
\begin{equation}
\label{eq:freq-zero}
\hat{X}^f_\ell[k] =
\begin{cases}
    0, & \text{if } k \in B_c, \\
    X^f_\ell[k], & \text{if } k \notin B_c.
\end{cases}
\end{equation}

Using these pruned coefficients $\hat{X}^f_\ell[k]$, we reconstruct the key and value tensors for layer $\ell$ via the IDFT. This reconstructed KV cache temporarily replaces the original one for that layer to evaluate the impact of pruning chunk $B_c$. We record the resulting model perplexity as $\mathrm{PPL}_{\ell,c}$. For baseline comparison, let $\mathrm{PPL}_{\mathrm{orig}}$ be the perplexity of the model with an unmodified KV cache. The normalized perplexity increase, $\Delta_{\ell,c}$, is then defined as:
\begin{equation}
\Delta_{\ell,c} = \frac{\mathrm{PPL}_{\ell,c} - \mathrm{PPL}_{\mathrm{orig}}}{\mathrm{PPL}_{\mathrm{orig}}},
\label{eq:delta-ppl}
\end{equation}
This metric, $\Delta_{\ell,c}$, quantifies the importance of frequency chunk $c$ at layer $\ell$; a larger value indicates a more significant contribution of that chunk to the model's performance.

Our analysis of these $\Delta_{\ell,c'}$ values reveals that while low-frequency components often demonstrate greater importance, many high-frequency components (or chunks containing them) also yield significant $\Delta$ values, indicating their non-negligible role. This finding suggests that a simple low-pass filtering approach might be suboptimal. The whole process is visualized in Appendix~\ref{fig:fabla}.

Thus, to retain as much critical information as possible across the spectrum, we employ a greedy compression strategy. For each layer, given a desired retention ratio $r$, we select the top $r \cdot C$ most important frequency chunks and discard the remainder. $C$ is a hyperparameter that controls the granularity of ablation study. We evaluate its effect in our experiments at Section~\ref{exp:abla}.

\section{FAEDKV}
\subsection{Infinite Window DFT}

The standard Discrete Fourier Transform (DFT), as defined in Equation~\eqref{eq:dft-full}, provides a static analysis of an entire input sequence. However, in autoregressive Transformer models, new tokens are generated sequentially. If we were to recompute the DFT over the entire growing KV cache (of current length $N$) at each decoding step, this would incur a computational cost of $O(N \log N)$ using FFT (or $O(N^2)$ naively) for that single step. This is problematic when compared to the typical complexities of decoder-only Transformers. We notice that \emph{sliding-window DFT} is better being recursively updated. With window size $M$, it updates each frequency bin as:
\begin{equation}
\label{eq:sliding-dft}
\begin{aligned}
    S_{t+1}[k] &= W_k \bigl(S_t[k] - x_{t-M+1} + x_{t+1}\bigr), \\
    W_k &= e^{-j\frac{2\pi k}{M}}, \quad \text{for } k=0, 1, \dots, M-1.
\end{aligned}
\end{equation}
Here, $x[t+1]$ is the new input sample entering the window, and $x[t-M+1]$ is the oldest sample leaving the window.The \emph{sliding-window DFT} efficiently updates the windowed frequency representation at $O(M)$ per time step to update all $M$ frequency bins.

Despite this, it presents two major drawbacks for our goal of KV cache compression. Firstly, it necessitates storing all $M$ time-domain samples of the current window. Secondly, this subtraction of $x[t-M+1]$ completely removes information about tokens older than the $M$-sample window.

% Start of the section discussing issues with simple infinite accumulation
To address the fixed memory window and associated storage overhead of the standard sliding-window DFT, an intuitive first step is to remove the subtraction of the oldest sample ($x[t-M+1]$). This creates a conceptually infinite, recursive window. However, this simpler recursion would lead to unbounded accumulation of values in the frequency-domain state $S_t[k]$, risking floating-point overflow for very long sequences common in LLMs\cite{lee2025infinitehipextendinglanguagemodel}.

Our Infinite Window DFT (IWDFT), defined in Equation~\ref{eq:infi-dft}, prevents such overflow by incorporating a normalization factor based on the current sequence length, $N$, into each recursive update:
\begin{equation}
\label{eq:infi-dft} % This is your existing label
S_{t+1}[k] = W_k \left( \frac{N-1}{N} S_t[k] + \frac{1}{N} x[t+1] \right).
\end{equation}
Here, $S_t[k]$ is the previous state for the $k$-th frequency bin. We approximate term $\frac{N-1}{N}$ to 1 since most context is larger than 1000.

This IWDFT approach is applied during decoding to update the K and V caches. It resolves the primary issues of the standard sliding window by avoiding extra time-domain storage and the hard cut-off of past information. Unlike methods that can introduce bias \cite{cai2024lococodroppingconvolutionslong,He_2023}, IWDFT processes each token's contribution consistently. The update is efficient ($O(N_{DFT})$ per step), training-free, and compatible with autoregressive LLMs.

\subsection{Workflow}
In this section, we detail our workflow, shown in Figure~\ref{fig:workflow}. Our overall idea is to transform Key and Value (KV) caches into the frequency domain using the Infinite Window DFT (IWDFT), and then compress this representation by selectively retaining frequency regions based on importance scores ($\Delta$) derived from an ablation study. Initially, for each model, this ablation study is performed.By measuring the normalized perplexity increase ($\Delta$) that occurs when each chunk is temporarily removed, we obtain layer-specific importance scores for these $C$ distinct spectral chunks. Given a desired retention ratio $r$, we then select the top $r \cdot C$ most important frequency chunks per layer based on these $\Delta$ scores. We refer to this process "pruning".

\subsubsection{Pre-filling Stage}
\label{subsec:prefilling_stage} % Optional label

In the pre-filling stage, an initial input context of length $N$ (e.g., up to $100k$ tokens) is processed with an approximate $O(N^2)$ attention complexity to generate the Key-Value (KV) cache. To prepare for compression, we transform a segment of this cache to the frequency domain using the Fast Fourier Transform (FFT) with Equation~\ref{eq:dft-full}). Guided by studies highlighting the importance of initial and recent tokens for "attention sinks"\cite{han2024lminfinitezeroshotextremelength}, we exclude the first $S$ and last $R$ tokens from this transformation. Thus, only the middle segment of $M=N-S-R$ tokens from each layer's KV cache undergoes DFT:
\begin{equation}
\label{eq:prefill-dft-succinct}
\begin{aligned}
    \mathbf{K}^f_{0:M-1} &= \mathrm{DFT}\bigl(\mathbf{K}[S:S+M-1]\bigr), \\
    \mathbf{V}^f_{0:M-1} &= \mathrm{DFT}\bigl(\mathbf{V}[S:S+M-1]\bigr).
\end{aligned}
\end{equation}
Subsequently, these $M$-length frequency-domain representations, $\mathbf{K}^f_{0:M-1}$ and $\mathbf{V}^f_{0:M-1}$, are pruned layer-wise. Using the set of important frequency components $B^*_\ell$ identified in Section~\ref{subsec:freq-ablation} and the rule from Equation~\ref{eq:freq-zero}, we obtain the compressed versions $\hat{\mathbf{K}}^f_{0:M-1}$ and $\hat{\mathbf{V}}^f_{0:M-1}$. Storing only these selected components significantly reduces memory for the full $N$-token KV cache from $O(N)$ to $O(N\cdot r)$, where $r$ denotes the cache compression ratio.

\subsubsection{Decoding Stage}
At each decoding step $t$, we have three operations: 

\textbf{Cache Reconstruction.} At each decoding step $t$, the compressed frequency-domain KV caches, $\hat{\mathbf{K}}^f_t$ and $\hat{\mathbf{V}}^f_t$ (inherited and updated from the previous step), are transformed back to the time domain using the IDFT operation from Equation~\ref{eq:inverse-swdft}. This yields the reconstructed caches $\mathbf{\widetilde{K}}_t$ and $\mathbf{\widetilde{V}}_t$:
\begin{equation}
\label{eq:reconstruct-idft-succinct}
\begin{aligned}
    \mathbf{\widetilde{K}}_t &= \mathrm{IDFT}\bigl(\hat{\mathbf{K}}^f_t\bigr), \\
    \mathbf{\widetilde{V}}_t &= \mathrm{IDFT}\bigl(\hat{\mathbf{V}}^f_t\bigr).
\end{aligned}
\end{equation}
To optimize this reconstruction, we leverage the sparsity inherent in the compressed $\hat{\mathbf{K}}^f_t$ and $\hat{\mathbf{V}}^f_t$ by employing a sparse IDFT implementation. It speed up the process by only working on the non-zero frequency components. We show this in our experiment at Section~\ref{exp:effi}

\textbf{Updating Cache}
As new tokens $(k_t, v_t)$ are generated and added to a recent time-domain window of size $R$, the tokens that age out of this window (e.g., $k_{t-R}, v_{t-R}$) are incorporated into the historical compressed KV cache. This update leverages our IWDFT mechanism, as defined in Equation~\ref{eq:infi-dft}. The update process is:
\begin{equation}
\label{eq:update-iwdft} % Changed label for clarity
\begin{aligned}
    \mathbf{K}^f_{t+1} &= \mathrm{IWDFT}\bigl(\hat{\mathbf{K}}^f_t, k_{t-R}\bigr), \\
    \mathbf{V}^f_{t+1} &= \mathrm{IWDFT}\bigl(\hat{\mathbf{V}}^f_t, v_{t-R}\bigr).
\end{aligned}
\end{equation}
Here, $\hat{\mathbf{K}}^f_t$ and $\hat{\mathbf{V}}^f_t$ are the previous compressed frequency-domain states, and $\mathbf{K}^f_{t+1}, \mathbf{V}^f_{t+1}$ are the updated states. This IWDFT process ensures that each token aging into the historical cache is incorporated consistently, allowing early context to maintain a sustained influence. Since the IWDFT update operates on all its maintained frequency bins, it can repopulate components that were previously zero due to pruning. After each IWDFT update, the pruning rule (Equation~\ref{eq:freq-zero}, using the selected components $B^*_\ell$) must be reapplied to maintain the compression level. In practice, we only compute and store the coefficients for the selected frequency components $B^*_\ell$.

\textbf{Calculating Attention} We assemble our $\mathbf{K}$ and $\mathbf{V}$ as follow:
\begin{equation}
 \label{eq:kv-concat}
 \begin{aligned}
 \mathbf{K}_t
 = \bigl[\,
 \mathbf{K}_{[0:S]},\;
 \mathbf{\widetilde{K}}_t,\;
 \mathbf{K}_{[N-R:N-1]}
 \bigr]
 ,\\
 \mathbf{V}_t
 = \bigl[\,
 \mathbf{V}_{[0:S]},\;
 \mathbf{\widetilde{K}}_t,\;
 \mathbf{V}_{[N-R:N-1]}
 \bigr],
 \end{aligned}
\end{equation}
As shown in the equation, the current KV pair consist of “attention‐sink” tokens, reconstructed tokens and the last $R$ recent tokens.

Finally, the attention output $\mathbf{o}^t$ is computed using the assembled $\mathbf{K}_t$ and $\mathbf{V}_t$ caches (Equation~\ref{eq:kv-attn}). Critically, FAEDKV requires no model fine-tuning for its integration. By reconstructing compressed KV cache segments to their original length, our method ensures the model operates within its existing architectural and positional embedding limits. This approach differs from other works\cite{tan2024llocolearninglongcontexts, cai2024lococodroppingconvolutionslong} that designed to extend the model's inherent context.

\section{Experiments}
\subsection{Setup}
\label{exp:setup}
We use Llama3-8B for long-context Question Answering (QA) and Qwen2-7B-Instruct for Needle-in-the-Haystack evaluations. Initial frequency ablation studies on both models informed our approach, leading us to segment the frequency spectrum into $C_{chunks}=22$ chunks (further details in Section~\ref{exp:abla}). Our method explicitly retains the first $S=10$ tokens as attention sinks and the most recent $R=50$ tokens, incurring negligible KV cache overhead from these. All experiments employed greedy decoding, and baseline methods were implemented using their officially provided code for fair comparison.

\subsection{QA Datasets}
\begin{table*}[t]
  \centering
  \resizebox{\textwidth}{!}{%
  \begin{tabular}{l
                  *{3}{c}  % Single‐Document QA
                  *{3}{c}  % Multi‐Document QA
                  *{3}{c}  % Summarization
                  *{3}{c}  % Few‐shot
                  *{2}{c}  % Synthetic
                  *{2}{c}  % Code
                  c        % Avg.
                  }
    \toprule
    \multirow{2}{*}{\textbf{Method}}
      & \multicolumn{3}{c}{\textbf{Single-Doc QA}}
      & \multicolumn{3}{c}{\textbf{Multi-Doc QA}}
      & \multicolumn{3}{c}{\textbf{Summarization}}
      & \multicolumn{3}{c}{\textbf{Few-shot}}
      & \multicolumn{2}{c}{\textbf{Synthetic}}
      & \multicolumn{2}{c}{\textbf{Code}}
      & \multirow{2}{*}{\textbf{Avg.}} \\
    \cmidrule(lr){2-4}  \cmidrule(lr){5-7}  \cmidrule(lr){8-10}
    \cmidrule(lr){11-13}  \cmidrule(lr){14-15}  \cmidrule(lr){16-17}
      & NtrvQA & QAsper & MF-en
      & HotpotQA & 2WikiMQA & Musique
      & GovReport & QMSum & MultiNews
      & TREC & TriviaQA & SAMSum
      & PCount & PRe
      & LCC & RB-P \\
    \midrule
    \textbf{FullKV} 
      & 22.53 & 26.29 & 41.52 
      & 37.46 & 29.24 & 21.56 
      & 29.47 & 22.23 & 25.85 
      & 65.00 & 81.00 & 40.12 
      & 4.65 & 4.75 
      & 39.42 & 43.63 
      & 33.42 \\
    \midrule
    \multicolumn{18}{c}{\textbf{Cache Size = 768 (Compression Ratio = 0.094)}} \\
    \midrule
    H2O     
      & 7.67 & 11.81 & 25.92 & 20.40 & 17.37 & 7.32 & 12.86 & 9.18 & 12.56 & 24.50 & 49.18 & 26.72 & 0.0 & 0.0 & 27.61 & 32.29 & 17.83 \\
    SnapKV  
      & 11.84 & 12.35 & 30.86 & 26.54 & 17.01 & 13.22 & 15.71 & 9.86 & 16.66 & 29.50 & 55.89 & 28.54 & 0 & 3.00 & 20.97 & 26.15 & 19.91\\
    FAEDKV   
       & 15.23 & 18.08 & 39.12 & 31.62 & 21.09 & 15.48 & 19.02 & 14.26 & 19.70 & 43.5 & 66.54 & 30.70 & 2.45 & 3.20 & 28.47 & 34.86 & \textbf{25.21}  \\
    \midrule
    \multicolumn{18}{c}{\textbf{Cache Size = 1024 (Compression Ratio = 0.125)}} \\
    \midrule
    H2O     
      & 16.97 & 19.94 & 41.54 & 34.26 & 24.03 & 17.53 & 22.83 & 19.40 & 23.26 & 55.50 & 78.65 & 33.68 & 3.24 & 4.25 & 33.89 & 38.52 & 28.69 \\
    SnapKV  
      & 17.98 & 19.35 & 40.67 & 35.02 & 24.46 & 16.24 & 24.72 & 18.00 & 23.27 & 60.00 & 77.73 & 35.03 & 3.18 & 4.16 & 33.07 & 38.62 & 28.86\\
    FAEDKV   
      & 17.49 & 20.24 & 42.40 & 34.26 & 25.84 & 17.39 & 23.13 & 18.55 & 21.61 & 61.50 & 77.51 & 34.27 & 3.26 & 4.50 & 35.30 & 39.68 & \textbf{29.45}\\
    \midrule
    \multicolumn{18}{c}{\textbf{Cache Size = 2048 (Compression Ratio = 0.25)}} \\
    \midrule
    H2O     
      & 21.63 & 23.89 & 45.32 
      & 37.53 & 29.21 & 20.72 
      & 28.11 & 22.54 & 25.78 
      & 63.50 & 81.10 & 38.47 
      & 3.50 & 4.50 
      & 38.13 & 43.22 
      & 32.95 \\
    SnapKV  
      & 22.12 & 25.26 & 44.03 
      & 38.13 & 30.06 & 22.99 
      & 28.53 & 22.53 & 26.35 
      & 64.50 & 80.10 & 38.57 
      & 3.50 & 4.50 
      & 32.19 & 38.22 
      & 32.60 \\
    FAEDKV   
      & 21.47 & 24.61 & 45.61 
      & 38.50 & 30.04 & 21.29 
      & 27.98 & 22.67 & 26.29 
      & 64.50 & 81.05 & 39.21 
      & 3.50 & 4.25 
      & 37.84 & 43.16 
      & \textbf{33.24} \\
    \bottomrule
  \end{tabular}%
  }
  \caption{LongBench performance (perplexity) across 16 tasks and varying cache sizes. Bold indicates the best compressed method at each cache size.}
  \label{tab:longbench_results}
\end{table*}

\label{exp:QA}
We evaluate our approach on Llama3-8B-Instruct model on the LongBench benchmark\cite{bai2024longbenchbilingualmultitaskbenchmark}, which comprises 16 tasks across six categories—single-document QA, multi-document QA, summarization, few-shot learning, synthetic tasks, and code completion—with an average context length of roughly 11 000 tokens. We compare against three baselines: \textbf{H2O}\cite{zhang2023h2oheavyhitteroracleefficient}, an eviction based compression method; SnapKV\cite{li2024snapkvllmknowslooking}, the state of art for long context tasks; and \textbf{full KV cache} that stores all KV pairs without compression. We set the baseline compression method's cache size to 512, 1024,2048 respectively. Correspondingly, we set our method's compression ratio $r$ to 0.094, 0.125 and 0.25 to facilitate fair comparison.

Table \ref{tab:longbench_results} presents our model’s performance across all LongBench tasks and cache sizes. On average, FAEDKV improves accuracy by 2.91 points compared to H2O and by 2.12 points compared to SnapKV, while it remains slightly below the full-attention baseline.

Importantly, under extremely tight cache budgets, FAEDKV outperforms eviction-based methods by up to \textbf{22 \%}. We attribute this advantage to the fact that eviction strategies tend to bias toward current tokens and discard valuable information that has low attention scores toward the current tokens. In contrast, our frequency-domain compression precisely identifies and preserves the most informative spectral components across the entire context, yielding a more balanced retention of information in the most constrained settings.

\begin{figure}[t]
  \includegraphics[width=1\columnwidth]{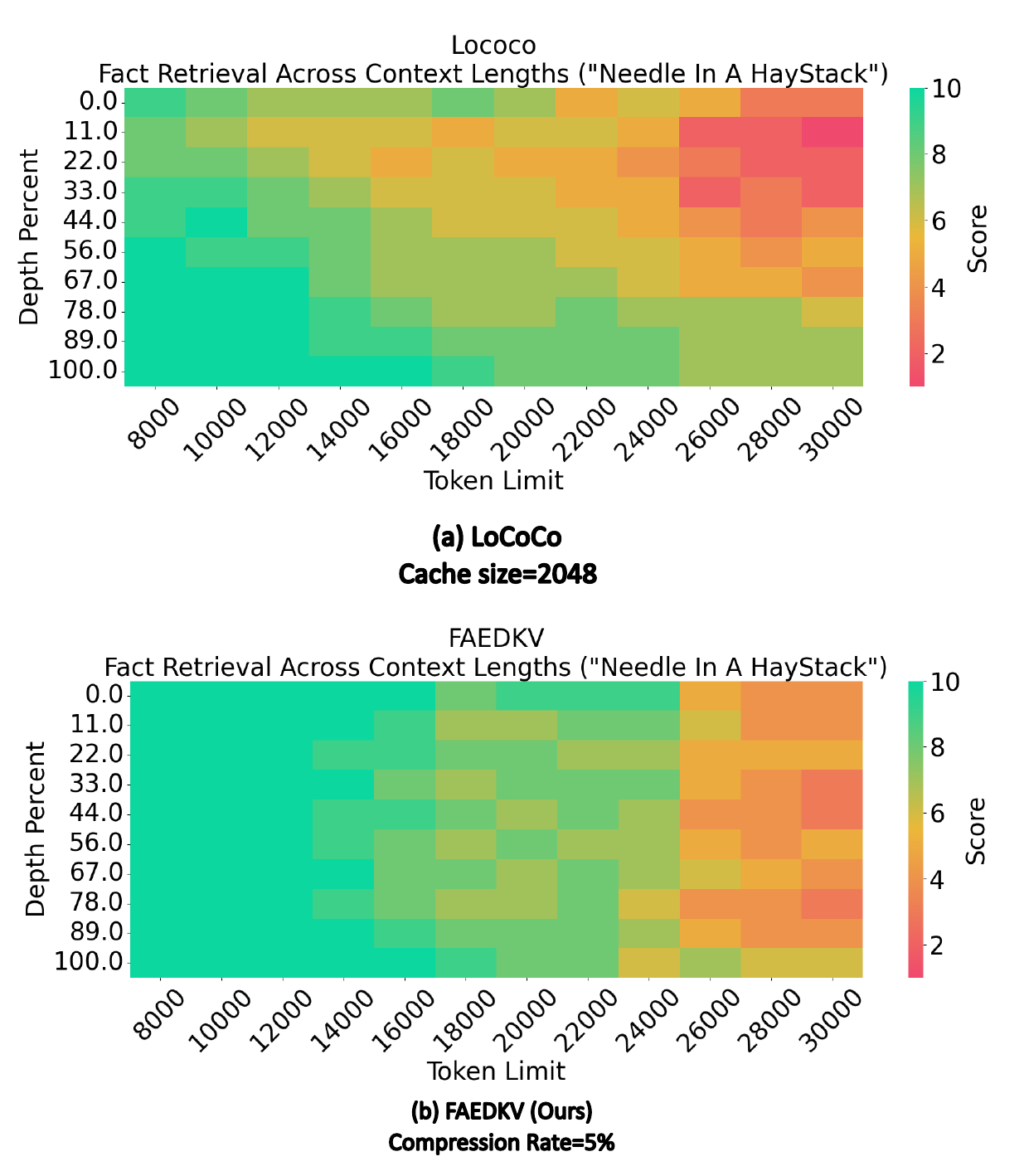}
  \caption{Results of Fact Retrieval Across Context Lengths (“Needle In A Haystack”).  The x-axis denotes the length
 of the document (the “haystack”) from 8K to 300K tokens; the axis indicates the position that the “needle” (a short sentence) is located within the document. }
  \label{fig:needle}
\end{figure}

\subsection{Fact Retrieval}
\label{exp:early}

To assess FAEDKV's ability to preserve information integrity across varying context lengths and token positions, we employed the "Needle in a Haystack" benchmark\cite{arxiv.2407.11963}. This test evaluates an LLM's capacity to retrieve specific information embedded within a larger text corpus. We used excerpts from THUDM's implementation\cite{bai-etal-2024-longalign}, creating contexts of 8K-30K tokens. A unique factual statement was inserted as the needle at 9 relative positions within each from 0\% to 100\% document depth.

Our experimental procedure involved presenting a Qwen2.5-7B-Instruct model with these augmented documents. We compared with LoCoCo\cite{cai2024lococodroppingconvolutionslong}, a convulution based compression approach. Both LoCoCo and FAEDKV were evaluated on 1024 ($r=0.05$ of 24K) Cache size.  

The results demonstrate two key advantages of FAEDKV. Firstly, FAEDKV generally achieved higher retrieval accuracy compared to the LoCoCo baseline across various context lengths and compression ratios. Secondly, and critically for our design, FAEDKV exhibited markedly more consistent accuracy irrespective of the needle's position within the haystack. This is attributed to its core mechanism employing the DFT, which inherently processes all token information with equal weight, ensuring a more uniform preservation of contextual details throughout the compressed KV cache.

\begin{figure}[t]
  \includegraphics[width=\columnwidth]{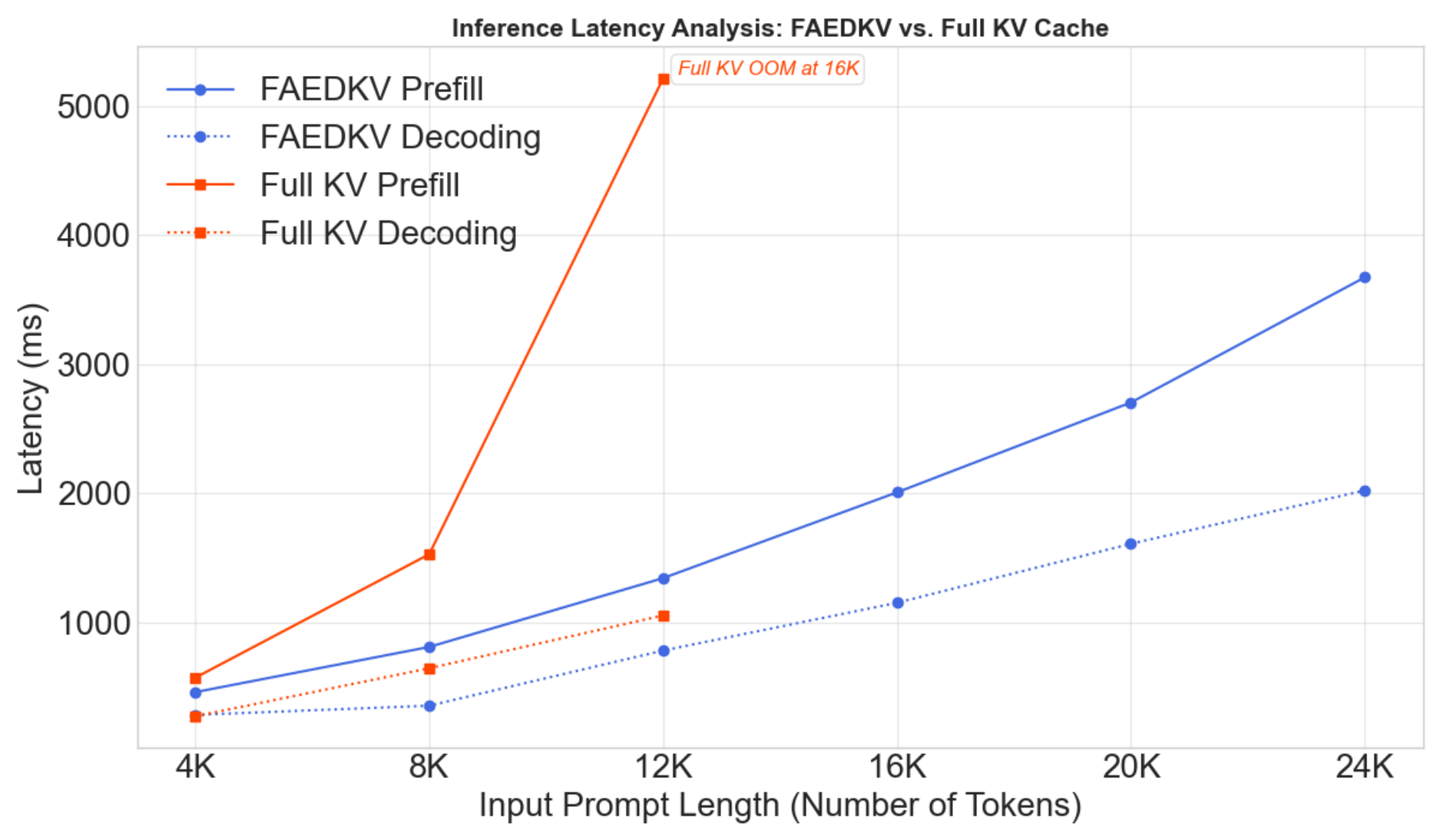}
  \caption{Results of Pre-filling and Decoding Latency}
  \label{fig:effi}
\end{figure}

\subsection{Pre-filling and Decoding Latency}
\label{exp:effi}
 To evaluate FAEDKV's computational efficiency, we benchmarked its inference latency against LoCoCo and a full KV cache baseline using a Llama-3 8B model (Hugging Face Transformers) on an NVIDIA A6000 GPU. We measured prefill latency and decoding latency for generating 10 tokens. We sample input prompts from the PG19\cite{raecompressive2019} dataset, pruned to lengths of 4K, 8K, 12K, and 16K tokens. All tests used a batch size of 1 and compression ratios of 0.1. As shown in Figure~\ref{fig:effi}, While the baseline exceeds GPU VRAM limits for sequences longer than 12k tokens, FAEDKV maintains efficient generation for sequences up to 24k tokens. Our result shows consistant improvement over the baseline, demonstrating the effectiveness of our optimizations in the workflow.

\subsection{Ablation Study Of Chunk Size}
\label{exp:abla}
\begin{figure}[t]
  \includegraphics[width=\columnwidth]{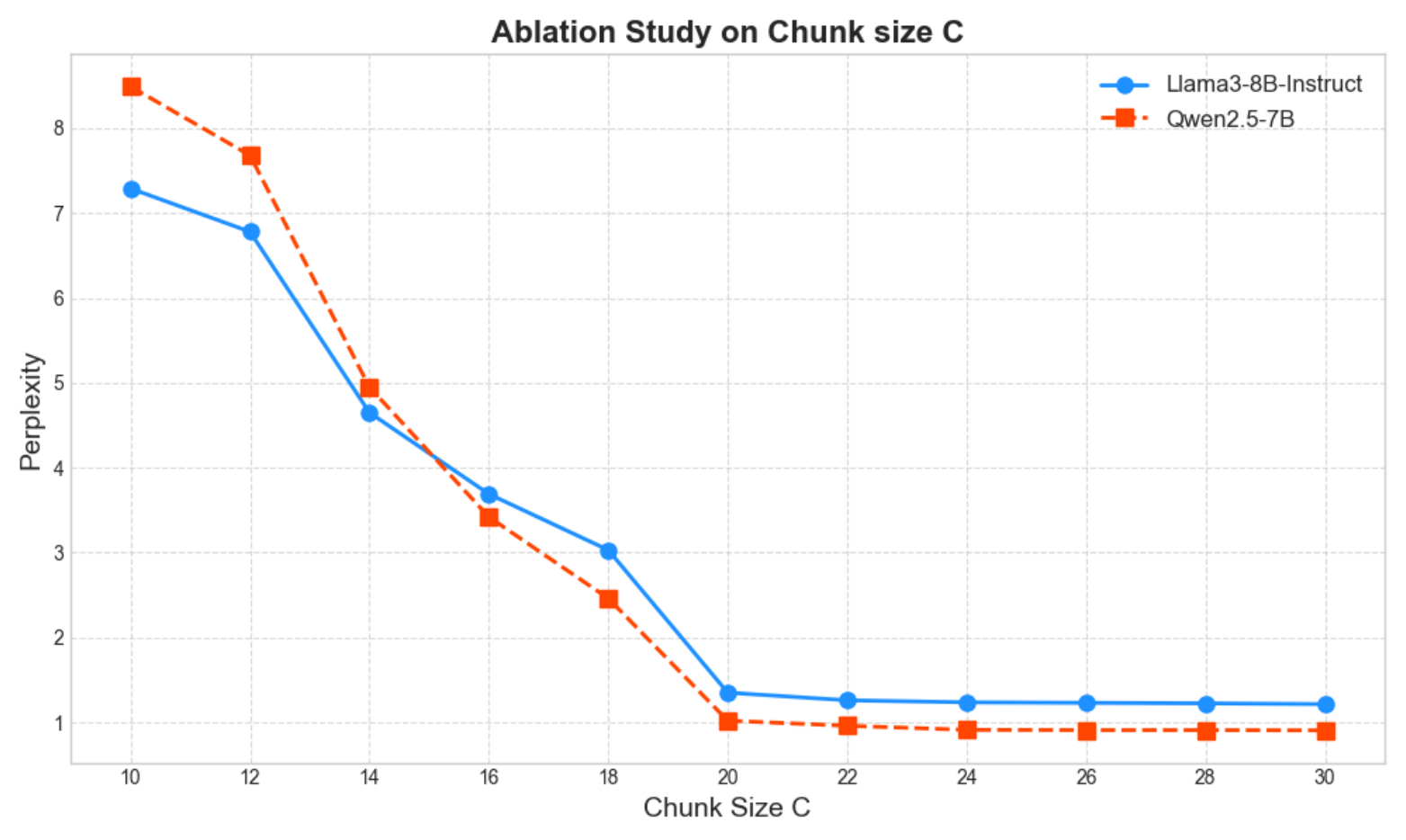}
  \caption{Results of Ablation Study on Chunk Size.}
  \label{fig:abla}
\end{figure}

We evaluated the impact of different chunk sizes ($C$)---a key hyperparameter for our frequency ablation study and subsequent workflow---on model performance. The ablation study was conducted with varying $C$ values, measuring perplexity increase on the PG-19 dataset for both Llama3-8B and Qwen2.5-7B models. As illustrated in Figure~\ref{fig:abla}, perplexity drops sharply around $C=12$ and stabilizes near its minimum at $C=22$. This finding supports $C=22$ as an optimal choice, balancing model performance retention with the computational cost of the ablation study.

\section{Conclusion}
\label{sec:conclusion}

In this paper, we introduced FAEDKV, a novel training-free KV cache compression algorithm designed to mitigate memory overhead in LLMs while promoting unbiased information retention. Our approach uniquely leverages frequency-domain transformations, guided by an empirical frequency ablation study to identify critical spectral components for preservation. The core of our method, the IWDFT, enables efficient and normalized updates to the compressed cache during autoregressive decoding, ensuring a more consistent treatment of token contributions over time. Experimental results on multiple benchmarks demonstrate FAEDKV's ability to achieve significant cache compression with competitive, and often superior, performance compared to existing methods, particularly in preserving information uniformly across long contexts.

\section{Limitations}

Our study has several limitations. Firstly, experiments were conducted on models deployable on a single A6000 GPU, simulating resource-constrained scenarios. While this provides practical insights, the behavior of significantly larger models in the frequency domain and the scalability of our approach warrant further investigation. Secondly, although FAEDKV improves efficiency, opportunities may exist for even more performant frequency-based inference by more deeply leveraging principles like signal locality or semantic clustering within the frequency domain. Finally, FAEDKV focuses on efficient KV cache management within a model's existing maximum context length and does not inherently extend this architectural limit,

\section{Ethical Considerations}

Large Language Models (LLMs), the systems our work aims to optimize, have well-documented broader ethical considerations. Our method, FAEDKV, is a technical contribution focused on improving computational efficiency via KV cache compression. As such, FAEDKV itself does not introduce new ethical dimensions beyond those inherent to the LLMs it is applied to, nor does it directly address these existing societal concerns.

\bibliography{custom}

\begin{thebibliography}{30}
\providecommand{\natexlab}[1]{#1}

\bibitem[{Ainslie et~al.(2023)Ainslie, Lee-Thorp, de~Jong, Zemlyanskiy, Lebron,
  and Sanghai}]{ainslie-etal-2023-gqa}
Joshua Ainslie, James Lee-Thorp, Michiel de~Jong, Yury Zemlyanskiy, Federico
  Lebron, and Sumit Sanghai. 2023.
\newblock \href {https://doi.org/10.18653/v1/2023.emnlp-main.298} {{GQA}:
  Training generalized multi-query transformer models from multi-head
  checkpoints}.
\newblock In \emph{Proceedings of the 2023 Conference on Empirical Methods in
  Natural Language Processing}, pages 4895--4901, Singapore. Association for
  Computational Linguistics.

\bibitem[{Bai et~al.(2024{\natexlab{a}})Bai, Lv, Zhang, He, Qi, Hou, Tang,
  Dong, and Li}]{bai-etal-2024-longalign}
Yushi Bai, Xin Lv, Jiajie Zhang, Yuze He, Ji~Qi, Lei Hou, Jie Tang, Yuxiao
  Dong, and Juanzi Li. 2024{\natexlab{a}}.
\newblock \href {https://doi.org/10.18653/v1/2024.findings-emnlp.74}
  {{L}ong{A}lign: A recipe for long context alignment of large language
  models}.
\newblock In \emph{Findings of the Association for Computational Linguistics:
  EMNLP 2024}, pages 1376--1395, Miami, Florida, USA. Association for
  Computational Linguistics.

\bibitem[{Bai et~al.(2024{\natexlab{b}})Bai, Lv, Zhang, Lyu, Tang, Huang, Du,
  Liu, Zeng, Hou, Dong, Tang, and
  Li}]{bai2024longbenchbilingualmultitaskbenchmark}
Yushi Bai, Xin Lv, Jiajie Zhang, Hongchang Lyu, Jiankai Tang, Zhidian Huang,
  Zhengxiao Du, Xiao Liu, Aohan Zeng, Lei Hou, Yuxiao Dong, Jie Tang, and
  Juanzi Li. 2024{\natexlab{b}}.
\newblock \href {https://arxiv.org/abs/2308.14508} {Longbench: A bilingual,
  multitask benchmark for long context understanding}.
\newblock \emph{Preprint}, arXiv:2308.14508.

\bibitem[{Cai et~al.(2024{\natexlab{a}})Cai, Tian, Wang, and
  Chen}]{cai2024lococodroppingconvolutionslong}
Ruisi Cai, Yuandong Tian, Zhangyang Wang, and Beidi Chen. 2024{\natexlab{a}}.
\newblock \href {https://arxiv.org/abs/2406.05317} {Lococo: Dropping in
  convolutions for long context compression}.
\newblock \emph{Preprint}, arXiv:2406.05317.

\bibitem[{Cai et~al.(2024{\natexlab{b}})Cai, Zhang, Gao, Liu, Liu, Lu, Xiong,
  Dong, Chang, Hu, and Xiao}]{cai2024pyramidkvdynamickvcache}
Zefan Cai, Yichi Zhang, Bofei Gao, Yuliang Liu, Tianyu Liu, Keming Lu, Wayne
  Xiong, Yue Dong, Baobao Chang, Junjie Hu, and Wen Xiao. 2024{\natexlab{b}}.
\newblock \href {https://arxiv.org/abs/2406.02069} {Pyramidkv: Dynamic kv cache
  compression based on pyramidal information funneling}.
\newblock \emph{Preprint}, arXiv:2406.02069.

\bibitem[{Ge et~al.(2023)Ge, Song, Liu, Liu, Wang, Wang, Zhou, Dou, and
  Wen}]{ge2023surveykvcache}
RangRang Ge, ShiYe Song, Zhaorui Liu, Wei Liu, Yuesheng Wang, Dongling Wang,
  Bofang Zhou, Zhicheng Dou, and Ji-Rong Wen. 2023.
\newblock A survey on {KV} cache compression for large language models.
\newblock \emph{arXiv preprint arXiv:2312.10546}.

\bibitem[{Ge et~al.(2024)Ge, Zhang, Liu, Zhang, Han, and
  Gao}]{ge2024modeltellsdiscardadaptive}
Suyu Ge, Yunan Zhang, Liyuan Liu, Minjia Zhang, Jiawei Han, and Jianfeng Gao.
  2024.
\newblock \href {https://arxiv.org/abs/2310.01801} {Model tells you what to
  discard: Adaptive kv cache compression for llms}.
\newblock \emph{Preprint}, arXiv:2310.01801.

\bibitem[{Han et~al.(2024)Han, Wang, Peng, Xiong, Chen, Ji, and
  Wang}]{han2024lminfinitezeroshotextremelength}
Chi Han, Qifan Wang, Hao Peng, Wenhan Xiong, Yu~Chen, Heng Ji, and Sinong Wang.
  2024.
\newblock \href {https://arxiv.org/abs/2308.16137} {Lm-infinite: Zero-shot
  extreme length generalization for large language models}.
\newblock \emph{Preprint}, arXiv:2308.16137.

\bibitem[{He et~al.(2023)He, Yang, Feng, Yin, Wang, Leng, and Lin}]{He_2023}
Ziwei He, Meng Yang, Minwei Feng, Jingcheng Yin, Xinbing Wang, Jingwen Leng,
  and Zhouhan Lin. 2023.
\newblock \href {https://doi.org/10.18653/v1/2023.findings-acl.570} {Fourier
  transformer: Fast long range modeling by removing sequence redundancy with
  fft operator}.
\newblock In \emph{Findings of the Association for Computational Linguistics:
  ACL 2023}, page 8954–8966. Association for Computational Linguistics.

\bibitem[{Kai et~al.(2025)Kai, Zeng, Wang, Bai, He, Jiang, and
  Lin}]{kai2025freqkvfrequencydomainkeyvalue}
Jushi Kai, Boyi Zeng, Yixuan Wang, Haoli Bai, Ziwei He, Bo~Jiang, and Zhouhan
  Lin. 2025.
\newblock \href {https://arxiv.org/abs/2505.00570} {Freqkv: Frequency domain
  key-value compression for efficient context window extension}.
\newblock \emph{Preprint}, arXiv:2505.00570.

\bibitem[{Lee et~al.(2024)Lee, Lee, Kim, and Heo}]{lee2024scaling}
Byungchan Lee, Seokmin Lee, Donghyun Kim, and Beomseok Heo. 2024.
\newblock Scaling {FFNs} for better transformer.
\newblock \emph{arXiv preprint arXiv:2403.15916}.

\bibitem[{Lee et~al.(2025)Lee, Park, Suh, and
  Hwang}]{lee2025infinitehipextendinglanguagemodel}
Heejun Lee, Geon Park, Jaduk Suh, and Sung~Ju Hwang. 2025.
\newblock \href {https://arxiv.org/abs/2502.08910} {Infinitehip: Extending
  language model context up to 3 million tokens on a single gpu}.
\newblock \emph{Preprint}, arXiv:2502.08910.

\bibitem[{Lee-Thorp et~al.(2021)Lee-Thorp, Ainslie, Eckstein, and
  Ontanon}]{lee2021fnet}
James Lee-Thorp, Joshua Ainslie, Ilya Eckstein, and Santiago Ontanon. 2021.
\newblock {FNet}: Mixing tokens with fourier transforms.
\newblock \emph{arXiv preprint arXiv:2105.03824}.

\bibitem[{Li et~al.(2024{\natexlab{a}})Li, Zhang, Zhang, Duan, Liu, and
  Chen}]{arxiv.2407.11963}
Mo~Li, Songyang Zhang, Taolin Zhang, Haodong Duan, Yunxin Liu, and Kai Chen.
  2024{\natexlab{a}}.
\newblock \href {https://doi.org/10.48550/ARXIV.2407.11963} {Needlebench: Can
  llms do retrieval and reasoning in information-dense context?}
\newblock \emph{Preprint}, arXiv:2407.11963.

\bibitem[{Li et~al.(2024{\natexlab{b}})Li, Ni, Li, Ma, and
  Wang}]{li2024afformer}
Shaoyi Li, Zhaowen Ni, Tiecheng Li, Hong Ma, and Zheng Wang.
  2024{\natexlab{b}}.
\newblock {AFFormer}: Resolution-agnostic and frequency-adaptive recurrent
  transformer for image super-resolution.
\newblock \emph{arXiv preprint arXiv:2403.05088}.

\bibitem[{Li et~al.(2024{\natexlab{c}})Li, Huang, Yang, Venkitesh, Locatelli,
  Ye, Cai, Lewis, and Chen}]{li2024snapkvllmknowslooking}
Yuhong Li, Yingbing Huang, Bowen Yang, Bharat Venkitesh, Acyr Locatelli,
  Hanchen Ye, Tianle Cai, Patrick Lewis, and Deming Chen. 2024{\natexlab{c}}.
\newblock \href {https://arxiv.org/abs/2404.14469} {Snapkv: Llm knows what you
  are looking for before generation}.
\newblock \emph{Preprint}, arXiv:2404.14469.

\bibitem[{Liu et~al.(2023{\natexlab{a}})Liu, Li, Chen, Li, and
  Wang}]{liu2023scissorhands}
Liyue Liu, Shijie Li, Zhuo Chen, Tianyi Li, and Yu~Wang. 2023{\natexlab{a}}.
\newblock \href {https://openreview.net/forum?id=aZIFC4NFfE} {Scissorhands:
  Exploiting the persistence of importance hypothesis for llm kv cache
  compression at test time}.
\newblock In \emph{Thirty-seventh Conference on Neural Information Processing
  Systems}.

\bibitem[{Liu et~al.(2023{\natexlab{b}})Liu, Lin, Hewitt, Paranjape,
  Bevilacqua, Petroni, and Liang}]{liu2023lostmiddlelanguagemodels}
Nelson~F. Liu, Kevin Lin, John Hewitt, Ashwin Paranjape, Michele Bevilacqua,
  Fabio Petroni, and Percy Liang. 2023{\natexlab{b}}.
\newblock \href {https://arxiv.org/abs/2307.03172} {Lost in the middle: How
  language models use long contexts}.
\newblock \emph{Preprint}, arXiv:2307.03172.

\bibitem[{Liu et~al.(2024)Liu, Wang, Zhao, Li, Bai, and Yu}]{liu2024deja}
Zichang Liu, Jue Wang, Tri Zhao, Zirui Li, Yixin Bai, and Jeff Yu. 2024.
\newblock Deja vu: Contextual sparsity for efficient {LLM} inference.
\newblock \emph{arXiv preprint arXiv:2401.09486}.

\bibitem[{Merity et~al.(2016)Merity, Xiong, Bradbury, and
  Socher}]{merity2016pointer}
Stephen Merity, Caiming Xiong, James Bradbury, and Richard Socher. 2016.
\newblock \href {https://arxiv.org/abs/1609.07843} {Pointer sentinel mixture
  models}.
\newblock \emph{Preprint}, arXiv:1609.07843.

\bibitem[{Murty et~al.(2024)Murty, Morris, Nanda, Li, Andreas, Hudson, and
  Misra}]{Murty2024ExactlyST}
Sharan Murty, John~D. Morris, Neel Nanda, Michael~J. Li, Jacob Andreas,
  Michael~I. Hudson, and Divya Misra. 2024.
\newblock Exactly solving acrostic puzzles with a language model.
\newblock \emph{arXiv preprint arXiv:2403.04534}.

\bibitem[{Nanda et~al.(2023)Nanda, Chan, Liberum, Smith, and
  Steinhardt}]{nanda2023emergent}
Neel Nanda, Lawrence Chan, Tom Liberum, Jess Smith, and Jacob Steinhardt. 2023.
\newblock \href {https://openreview.net/forum?id=zNHz3V8xS5} {Emergent linear
  representations in world models of self-supervised transformers}.
\newblock In \emph{International Conference on Learning Representations}.

\bibitem[{Rae et~al.(2019)Rae, Potapenko, Jayakumar, Hillier, and
  Lillicrap}]{raecompressive2019}
Jack~W Rae, Anna Potapenko, Siddhant~M Jayakumar, Chloe Hillier, and Timothy~P
  Lillicrap. 2019.
\newblock \href {https://arxiv.org/abs/1911.05507} {Compressive transformers
  for long-range sequence modelling}.
\newblock \emph{arXiv preprint}.

\bibitem[{Su et~al.(2024)Su, Lu, Pan, Murtadha, Wen, and Liu}]{su2024roformer}
Jianlin Su, Yu~Lu, Shengfeng Pan, Ahmed Murtadha, Bo~Wen, and Yunfeng Liu.
  2024.
\newblock {RoFormer}: Enhanced transformer with rotary position embedding.
\newblock \emph{arXiv preprint arXiv:2104.09864}.

\bibitem[{Tan et~al.(2024)Tan, Li, Patil, Wu, Zhang, Keutzer, Gonzalez, and
  Popa}]{tan2024llocolearninglongcontexts}
Sijun Tan, Xiuyu Li, Shishir Patil, Ziyang Wu, Tianjun Zhang, Kurt Keutzer,
  Joseph~E. Gonzalez, and Raluca~Ada Popa. 2024.
\newblock \href {https://arxiv.org/abs/2404.07979} {Lloco: Learning long
  contexts offline}.
\newblock \emph{Preprint}, arXiv:2404.07979.

\bibitem[{Vaswani et~al.(2023)Vaswani, Shazeer, Parmar, Uszkoreit, Jones,
  Gomez, Kaiser, and Polosukhin}]{vaswani2023attentionneed}
Ashish Vaswani, Noam Shazeer, Niki Parmar, Jakob Uszkoreit, Llion Jones,
  Aidan~N. Gomez, Lukasz Kaiser, and Illia Polosukhin. 2023.
\newblock \href {https://arxiv.org/abs/1706.03762} {Attention is all you need}.
\newblock \emph{Preprint}, arXiv:1706.03762.

\bibitem[{Zhang et~al.(2024)Zhang, Liu, Xiao, Shao, Ye, and
  Dou}]{zhang2024longcontextcompressionactivation}
Peitian Zhang, Zheng Liu, Shitao Xiao, Ninglu Shao, Qiwei Ye, and Zhicheng Dou.
  2024.
\newblock \href {https://arxiv.org/abs/2401.03462} {Long context compression
  with activation beacon}.
\newblock \emph{Preprint}, arXiv:2401.03462.

\bibitem[{Zhang et~al.(2023{\natexlab{a}})Zhang, Sheng, He, Tan, Zhou, Meng,
  Yu, Zhao, Chen, Su et~al.}]{zhang2023h2o}
Zhenyu Zhang, Ying Sheng, Tianyi He, Chen Tan, Yuesong Zhou, Lian Meng, Kai Yu,
  Aston Zhao, Haotong Chen, Jiaming Su, and 1 others. 2023{\natexlab{a}}.
\newblock H2o: Heavy-hitter oracle for efficient generative inference of large
  language models.
\newblock \emph{arXiv preprint arXiv:2306.14048}.

\bibitem[{Zhang et~al.(2023{\natexlab{b}})Zhang, Sheng, Zhou, Chen, Zheng, Cai,
  Song, Tian, Ré, Barrett, Wang, and
  Chen}]{zhang2023h2oheavyhitteroracleefficient}
Zhenyu Zhang, Ying Sheng, Tianyi Zhou, Tianlong Chen, Lianmin Zheng, Ruisi Cai,
  Zhao Song, Yuandong Tian, Christopher Ré, Clark Barrett, Zhangyang Wang, and
  Beidi Chen. 2023{\natexlab{b}}.
\newblock \href {https://arxiv.org/abs/2306.14048} {H$_2$o: Heavy-hitter oracle
  for efficient generative inference of large language models}.
\newblock \emph{Preprint}, arXiv:2306.14048.

\bibitem[{Zhou et~al.(2024)Zhou, Wang, Yin, Zhou, Cao, Gao, and
  Zhao}]{zhou2024sfm}
Kaixuan Zhou, Jianing Wang, Zhiyuan Yin, Yan Zhou, Xin Cao, Yang Gao, and Sheng
  Zhao. 2024.
\newblock {SFM-LLM}: A efficient long-term time series forecasting framework
  based on spectrum frequency mix.
\newblock \emph{arXiv preprint arXiv:2403.15912}.

\end{thebibliography}

\appendix
\begin{figure*}[t]
  \includegraphics[width=2\columnwidth]{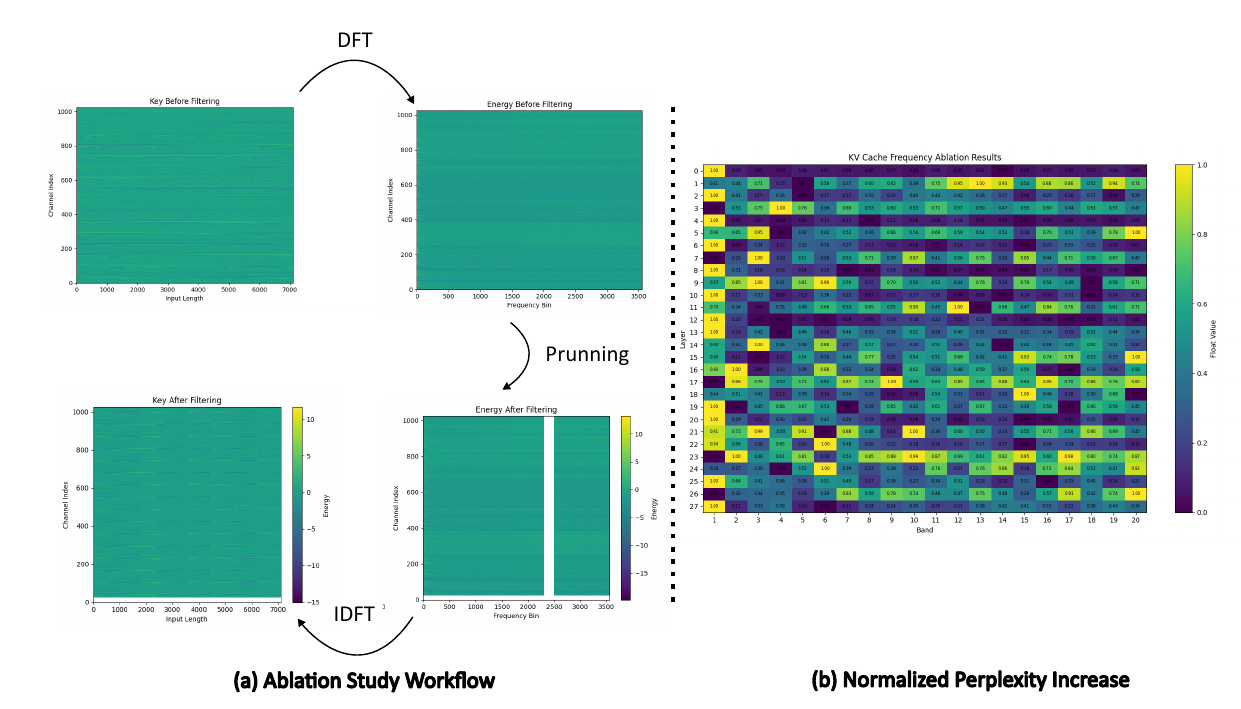}
  \caption{Overview of the Frequency Ablation Study. 
    (a) The workflow illustrates the process: time-domain Key/Value cache data (``Key Before Filtering'') is transformed via DFT into its frequency-domain energy representation (``Energy Before Filtering'')[cite: 1]. Specific frequency bands are then pruned (``Energy After Filtering''), and the corresponding time-domain data is reconstructed via IDFT (``Key After Filtering'') to evaluate performance impact[cite: 1]. 
    (b) A heatmap displays the Normalized Perplexity Increase ($\Delta_{\ell,c}$) resulting from ablating different frequency bands (x-axis, 1-20) across various model layers (y-axis, 1-27)[cite: 1]. Higher float values (brighter colors, from 0.0 to 1.0) indicate greater importance of the ablated band to model performance[cite: 1].}
  \label{fig:fabla}
\end{figure*}

\end{document}